\documentclass[runningheads]{llncs}
\usepackage{graphicx}
\usepackage[table,xcdraw]{xcolor}
\usepackage{amsfonts}
\usepackage{amsmath}
\usepackage{hyperref}
\usepackage{ amssymb }
\usepackage{graphicx}
\usepackage{soul,color}
\usepackage{chapterbib}
\usepackage{ amssymb }
\usepackage{graphicx}
\usepackage{soul,color}
\usepackage{multirow}
\usepackage{hyperref}
\begin{document}
	\title{Medical Transformer: Gated Axial-Attention for Medical Image Segmentation }
	\titlerunning{Medical Transformer}
	%
	\author{Jeya Maria Jose Valanarasu\inst{1} \and 
		Poojan Oza\inst{1} \and
		Ilker Hacihaliloglu\inst{2} \and
		Vishal M. Patel\inst{1}
	}
	%
	\authorrunning{JMJ Valanarasu et al.}
	%
	\institute{Johns Hopkins University, Baltimore, MD, USA \and
		Rutgers, The State University of New Jersey, NJ, USA 
	}
	\maketitle              
	\begin{abstract}
		
		Over the past decade, deep convolutional neural networks have been widely adopted for medical image segmentation and shown to achieve adequate performance. However, due to inherent inductive biases present in convolutional architectures, they lack understanding of long-range dependencies in the image. Recently proposed transformer-based architectures that leverage self-attention mechanism  encode long-range dependencies and learn representations that are highly expressive. This motivates us to explore transformer-based solutions and study the feasibility of using transformer-based network architectures for medical image segmentation tasks. Majority of existing transformer-based network architectures proposed for vision applications  require large-scale datasets to train properly. However, compared to the datasets for vision applications, in medical imaging the number of data samples is relatively low, making it difficult to efficiently train transformers for medical imaging  applications. To this end, we propose a gated axial-attention model which extends the existing architectures by introducing an additional control mechanism in the self-attention module. Furthermore, to train the model effectively on medical images, we propose a Local-Global training strategy (LoGo) which further improves the performance. Specifically, we operate on the whole image and patches to learn global and local features, respectively. The proposed Medical Transformer (MedT) is evaluated on three different medical image segmentation datasets and it is shown that it achieves better performance than the convolutional and other related transformer-based architectures. Code: \href{https://github.com/jeya-maria-jose/Medical-Transformer}{https://github.com/jeya-maria-jose/Medical-Transformer}
		
	\end{abstract}

	\section{Introduction}
	 
		Developing automatic, accurate, and robust medical image segmentation methods have been one of the principal problems in medical imaging as it is essential for computer-aided diagnosis and image-guided surgery systems. Segmentation of organs or lesion from a medical scan helps clinicians make an accurate diagnosis, plan the surgical procedure, and propose treatment strategies. Following the popularity of deep convolutional neural networks (ConvNets) in computer vision, ConvNets were quickly adopted for medical image segmentation. Networks like U-Net \cite{ronneberger2015u}, V-Net \cite{milletari2016v}, 3D U-Net \cite{cciccek20163d}, Res-UNet \cite{xiao2018weighted}, Dense-UNet \cite{li2018h}, Y-Net \cite{mehta2018net}, U-Net++ \cite{zhou2018unet++}, KiU-Net \cite{valanarasu2020kiu,valanarasu2020kiu1} and U-Net3+ \cite{huang2020unet} have been proposed specifically for performing image and volumetric segmentation for various medical imaging modalities. These methods achieve impressive performance on many difficult datasets, proving the effectiveness of ConvNets in learning discriminative features to segment the organ or lesion from a medical scan.
	 
		ConvNets are currently the basic building blocks of most methods proposed for image segmentation. However, they lack the ability to model long-range dependencies present in an image. More precisely, in ConvNets each convolutional kernel attends to only a local-subset of pixels in the whole image and forces the network to focus on local patterns rather than the global context. There have been works that have focused on modeling long-range dependencies for ConvNets using image pyramids \cite{zhao2017pyramid}, atrous convolutions \cite{chen2014semantic} and attention mechanisms \cite{huang2019ccnet}. However, it can be noted that there is still a scope of improvement for modeling long-range dependencies as the majority of previous methods do not focus on this aspect for medical image segmentation tasks.

	 \begin{figure}[htbp]
	 	\vskip -20pt
	 	\centering
	 	\includegraphics[width=0.75\linewidth]{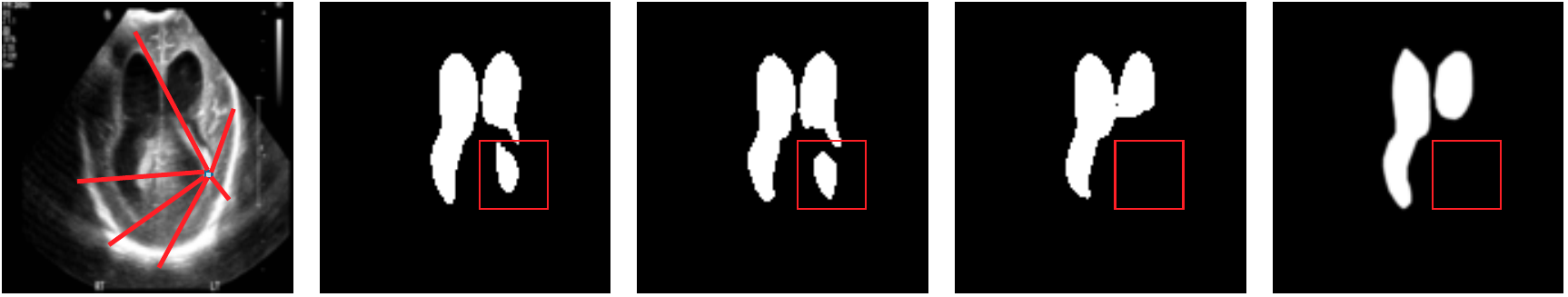}\\
	 	(a) \hskip35pt (b) \hskip35pt (c) \hskip35pt (d) \hskip35pt (e) \hskip35pt 
	 	\vskip -10.0pt
	 	\caption{(a) Input Ultrasound of in vivo preterm neonatal brain ventricle. Predictions by (b) U-Net, (c) Res-UNet, (d) MedT, and (e) Ground Truth. The red box highlights the region which are miss-classified by ConvNet based methods due to lack of learned long-range dependencies. The ground truth here was segmented by an expert clinician. Although it shows some bleeding inside the ventricle area, it does not correspond to the segmented area. This information is correctly captured by transformer-based models. }
	 	\label{explana}
	 \end{figure}

	To first understand why long-range dependencies matter for medical images, we visualize an example ultrasound scan of a preterm neonate and segmentation predictions of brain ventricles from the scan in Fig \ref{explana}. For a network to provide an efficient segmentation, it should be able to understand which pixels correspond to the mask and which to the background. As the background of the image is scattered, learning long-range dependencies between the pixels corresponding to the background can help in the network to prevent miss-classifying a pixel as the mask leading to reduction of false positives (considering 0 as background and 1 as segmentation mask). Similarly, whenever the segmentation mask is large, learning long-range dependencies between the pixels corresponding to the mask is also helpful in making efficient predictions. In Fig \ref{explana} (b) and (c), we can see that the convolutional networks miss-classify the background as a brain ventricle while the proposed transformer-based method does not make that mistake. This happens as our proposed method learns long-range dependencies of the pixel regions with that of the background.


	 In many natural language processing (NLP) applications, transformers \cite{devlin2018bert} have shown to be able to encode long-range dependencies. This is due to the self-attention mechanism which finds the dependency between given sequential input. Following their popularity in NLP applications, transformers have been adopted to computer vision applications very recently \cite{dosovitskiy2020image,touvron2020training}. With regard to transformers for segmentation tasks, Axial-Deeplab \cite{wang2020axial} utilized the axial attention module \cite{ho2019axial}, which factorizes 2D self-attention into two 1D self-attentions and introduced position-sensitive axial attention design for segmentation. In Segmentation Transformer (SETR) \cite{zheng2020rethinking}, a transformer was used as encoder which inputs a sequence of image patches and a ConvNet was used as decoder resulting in a powerful segmentation model. In medical image segmentation, transformer-based models have not been explored much. The closest works are the ones that use attention mechanisms to boost the performance  \cite{oktay2018attention,wang2019volumetric}. However, the encoder and decoder of these networks still have convolutional layers as the main building blocks.  
	 
	 It was observed that that the transformer-based models work well only when they are trained on large-scale datasets \cite{dosovitskiy2020image}. This becomes problematic while adopting transformers for medical imaging tasks as the number of images, with corresponding labels, available for training in any medical dataset is relatively scarce. Labeling process is also expensive and requires expert knowledge. Specifically, training with fewer images causes difficulty in learning positional encoding for the images. To this end, we propose a gated position-sensitive axial attention mechanism where we introduce four gates that control the amount of information the positional embedding supply to key, query, and value. These gates are learnable parameters which make the proposed mechanism to be applied to any dataset of any size. Depending on the size of the dataset, these gates would learn whether the number of images would be sufficient enough to learn proper position embedding. Based on whether the information learned by the positional embedding is useful or not, the gate parameters either converge to 0 or to some higher value. Furthermore, we propose a Local-Global (LoGo) training strategy, where we use a shallow global branch and a deep local branch that operates on the patches of the medical image. This strategy improves the segmentation performance  as we do not only operate on the entire image but focus on finer details present in the local patches. Finally, we propose Medical Transformer (MedT), which uses our gated position-sensitive axial attention as the building blocks and adopts our LoGo training strategy.
	 
	 In summary, this paper (1) proposes a gated position-sensitive axial attention mechanism that works well even on smaller datasets, (2) introduces Local-Global (LoGo) training methodology for transformers which is effective, (3) proposes Medical-Transformer (MedT) which is built upon the above two concepts proposed specifically for medical image segmentation, and (4) successfully improves the performance for medical image segmentation tasks over convolutional networks and fully attention architectures on three different datasets.

	\section{Medical Transformer (MedT)}
	


		
	\begin{figure}[b!]
		\centering
		\includegraphics[width=0.9\linewidth]{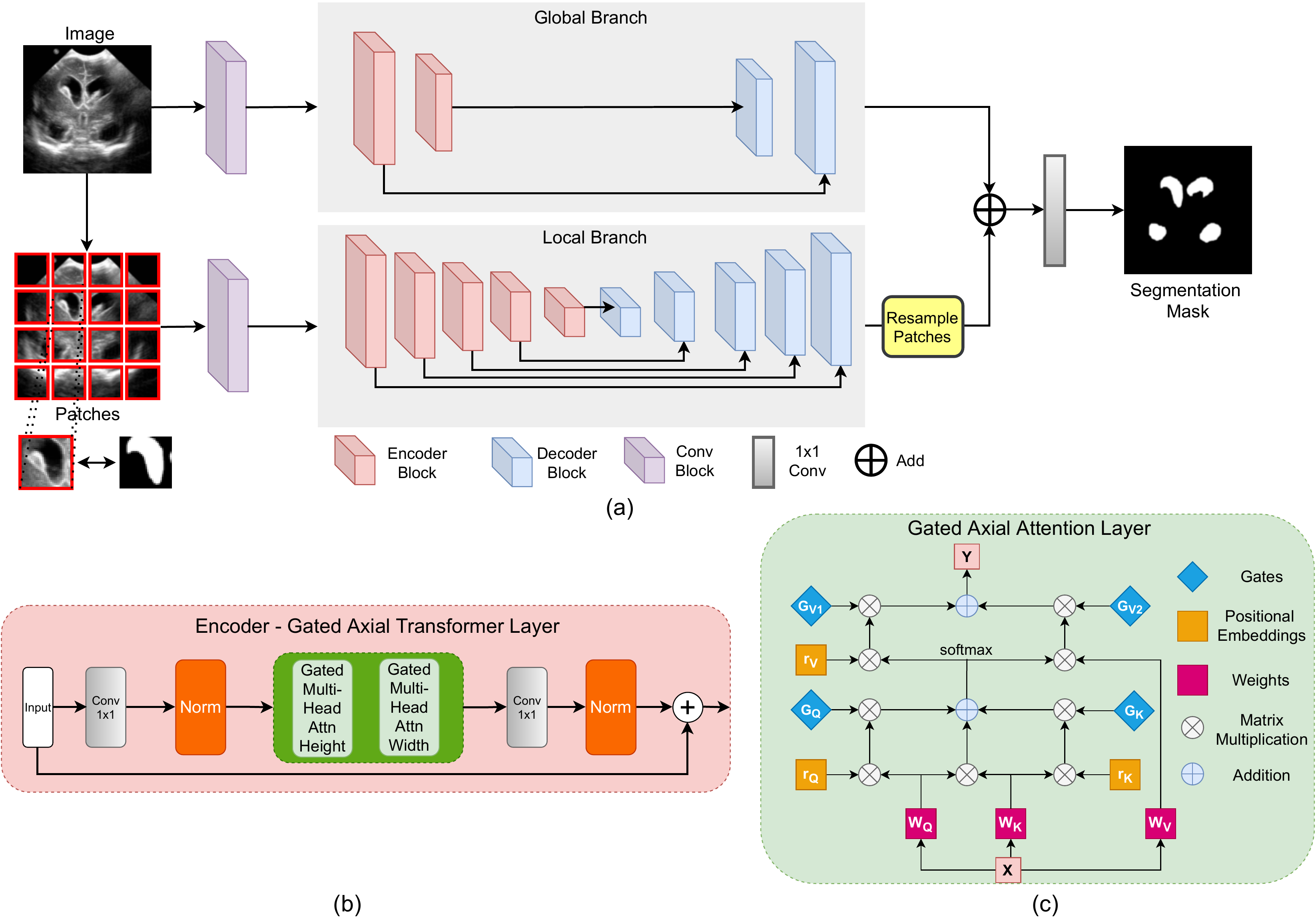}\\
		\vskip -12.5pt
		\caption{ (a) The main architecture diagram of MedT which uses LoGo strategy for training. (b) The gated axial transformer layer which is used in MedT. (c) Gated Axial Attention layer which is the basic building block of both height and width gated multi-head attention blocks found in the gated axial transformer layer. }
		
		\label{medt}
	\end{figure}

	\subsection{Self-Attention Overview}
	%
	
	Let us consider an input feature map $x \in \mathbb{R}^{C_{in} \times H \times W}$ with height $H$, weight $W$ and channels $C_{in}$. The output $y \in \mathbb{R}^{C_{out} \times H \times W}$ of a self-attention layer is computed with the help of projected input using the following equation:
	\begin{equation}\label{eq:self_attention}
	y_{ij} \ = \ \sum_{h=1}^{H} \sum_{w=1}^{W} \operatorname{softmax} \left(q_{ij}^{T} k_{hw}\right) v_{hw},
	\end{equation}
	where queries $q=W_Q x$, keys $k=W_K x$ and values $v=W_V x$ are all projections computed from the input $x$. Here, $q_{ij}, k_{ij}, v_{ij}$ denote query, key and value at any arbitrary location $i \in \{1, \dots, H\}$ and $j \in \{1, \dots, W\}$, respectively. The projection matrices $W_Q, W_K, W_V \in \mathbb{R}^{C_{in} \times C_{out}}$ are learnable. As shown in Eq.~\ref{eq:self_attention}, the values $v$ are pooled based on global affinities calculated using $\operatorname{softmax} (q^T k)$. Hence, unlike convolutions the self-attention mechanism is able to capture non-local information from the entire feature map. However, computing such affinities are computationally very expensive and with increased feature map size it often becomes infeasible to use self-attention for vision model architectures. Moreover, unlike convolutional layer, self-attention layer does not utilize any positional information while computing the non-local context. Positional information is often useful in vision models to capture structure of an object.
	
	\subsubsection{Axial-Attention}
	To overcome the computational complexity of calculating the affinities, self-attention is decomposed into two self-attention modules. The first module performs self-attention on the feature map height axis and the second one operates on the width axis. This is referred to as axial attention \cite{ho2019axial}. The axial attention consequently applied on height and width axis effectively model original self-attention mechanism with much better computational efficacy. To add positional bias while computing affinities through self-attention mechanism, a position bias term is added to make the affinities sensitive to the positional information \cite{shaw2018self}. This bias term is often referred to as relative positional encodings. These positional encodings are typically learnable through training and have been shown to have the capacity to encode spatial structure of the image. Wang \emph{et al.} \cite{wang2020axial} combined both the axial-attention mechanism and positional encodings to propose an attention-based model for image segmentation. Additionally, unlike previous attention model which utilizes relative positional encodings only for queries, Wang \emph{et al.} \cite{wang2020axial} proposed to use it for all queries, keys and values. This additional position bias in query, key and value is shown to capture long-range interaction with precise positional information \cite{wang2020axial}. For any given input feature map $x$, the updated self-attention mechanism with positional encodings along with width axis can be written as:
	\begin{equation}\label{eq:axial_deeplab}
	y_{ij} \ = \ \sum_{w=1}^{W} \operatorname{softmax} \left(q_{ij}^{T} k_{iw} + q_{ij}^{T} r^q_{iw} + k_{iw}^{T} r^k_{iw} \right) (v_{iw} + r^v_{iw}),
	\end{equation}
	where the formulation in Eq.~\ref{eq:axial_deeplab} follows the attention model proposed in \cite{wang2020axial} and $r^q, r^k, r^v \in \mathbb{R}^{W \times W}$ for the width-wise axial attention model. Note that  Eq.~\ref{eq:axial_deeplab} describes the axial attention applied along  the width axis of the tensor.   A similar formulation is also used to apply axial attention along the height axis and together they form a single self-attention model that is computationally efficient.
	
	\subsection{Gated Axial-Attention} We discussed the benefits of using the axial-attention mechanism proposed in \cite{wang2020axial} for visual recognition. Specifically, the axial-attention proposed in \cite{wang2020axial} is able to compute non-local context with good computational efficiency, able to encode positional bias into the mechanism and enables the ability to encode long-range interaction within an input feature map. However, their model is evaluated on large-scale segmentation datasets and hence it is easier for the axial-attention to learn positional bias at key, query and value. We argue that for experiments with small-scale datasets, which is often the case in medical image segmentation, the positional bias is difficult to learn and hence will not always be accurate in encoding long-range interactions. In the case where the learned relative positional encodings are not accurate enough, adding them to the respective key, query and value tensor would result in reduced performance. Hence, we propose a modified axial-attention block that can control the influence positional bias can exert in the encoding of non-local context. With the proposed modification the self-attention mechanism applied on the width axis can be formally written as:
	\begin{equation}\label{eq:gated_axial}
	y_{ij} \ = \ \sum_{w=1}^{W} \operatorname{softmax} \left(q_{ij}^{T} k_{iw} + G_Q q_{ij}^{T} r^q_{iw} + G_K k_{iw}^{T} r^k_{iw}\right) ( G_{V1} v_{iw} + G_{V2} r^v_{iw}),
	\end{equation}
	where the self-attention formula closely follows Eq.~\ref{eq:axial_deeplab} with added gating mechanism. Also, $G_Q, G_K, G_{V1}, G_{V2} \in \mathbb{R}$ are learnable parameters and together they create gating mechanism which control influence of the learned relative positional encodings have on encoding non-local context. Typically, if a relative positional encoding is learned accurately, the gating mechanism will assign it high weight compared to the ones which are not learned accurately. Fig ~\ref{medt} (c) illustrates the feed-forward in a typical gated axial attention layer.
	
	\subsection{Local-Global Training}
	
	It is evident that a transformer on patches is faster but patch-wise training alone is not sufficient for the tasks like medical image segmentation. Patch-wise training restricts the network in learning any information or dependencies for inter-patch pixels. To improve the overall understanding of the image, we propose to use two branches in the network, i.e., a global branch which works on the original resolution of the image, and a local branch which operates on patches of the image.	In the global branch, we reduce the number of gated axial transformer layers as we observe that the first few blocks of the proposed transformer model is sufficient to model long range dependencies. In the local branch, we create 16 patches of size $I/4 \times I/4$ of the image where $I$ is the dimensions of the original image. In the local branches, each patch is feed forwarded through the network and the output feature maps are re-sampled based on their  location to get the output feature maps. The output feature maps of both of the branches are then added and passed through a $1 \times 1$ convolution layer to produce the output segmentation mask. This strategy improves the performance as the global branch focuses on high-level information and the local branch can focus on finer details. The proposed Medical Transformer (MedT) uses gated axial attention layer as the basic building block and uses LoGo strategy for training. It is illustrated in Fig \ref{medt} (a). More details on the architecture and an ablation study with regard to the architecture can be found in the supplementary file.
	
	\section{Experiments and Results}
	
	\subsection{Dataset details}
	
	We use Brain anatomy segmentation (ultrasound) \cite{wang2018automatic,valanarasu2020learning}, Gland segmentation (microscopic) \cite{sirinukunwattana2017gland} and MoNuSeg (microscopic) \cite{kumar2019multi,kumar2017dataset} datasets for evaluating our method. More details about the datasets can be found in the supplementary.
	
	
	\subsection{Implementation details}
	We use binary cross-entropy (CE) loss between the prediction and the ground truth to train our network and can be written as:	
	\[\mathcal{L}_{CE(p,\hat{p})} = - \left(\frac{1}{wh} \sum_{x=0}^{w-1}\sum_{y=0}^{h-1}(p(x,y) \log(\hat{p}(x,y)) ) + (1-p(x,y))\log(1-\hat{p}(x,y))\right)\]	
	where $w$ and $h$ are the dimensions of the image,  $p(x,y)$ corresponds to the pixel in the image and  $\hat{p}(x,y)$ denotes the output prediction at a specific location $(x,y)$. The training details are provided in the supplementary document.
	
	For baseline comparisons, we first run experiments on both convolutional and transformer-based methods. For convolutional baselines, we compare with fully convolutional network (FCN) \cite{badrinarayanan2017segnet}, U-Net \cite{ronneberger2015u}, U-Net++ \cite{zhou2018unet++} and Res-Unet \cite{xiao2018weighted}. For transformer-based baselines, we use Axial-Attention U-Net with residual connections inspired from \cite{wang2020axial}. For our proposed method, we experiment with all the individual contributions. In gated axial attention network, we use axial attention U-Net with all its axial attention layers replaced with the proposed gated axial attention layers. In LoGo, we perform local global training for axial attention U-Net without using the gated axial attention layers. In MedT, we use gated axial attention as the basic building block for global branch and axial attention without positional encoding for local branch.
	
	\subsection{Results}
	
	\begin{table}[]
		\vskip -25pt 
		\centering
		\caption{Quantitative comparison of the proposed methods with convolutional and transformer based baselines in terms of F1 and IoU scores.}
		\label{table:res}
		\vskip -7.5pt
		\resizebox{0.75\linewidth}{!}{
		\begin{tabular}{
				>{\columncolor[HTML]{FFFFFF}}c |
				>{\columncolor[HTML]{FFFFFF}}c |
				>{\columncolor[HTML]{FFFFFF}}c 
				>{\columncolor[HTML]{FFFFFF}}c |
				>{\columncolor[HTML]{FFFFFF}}c 
				>{\columncolor[HTML]{FFFFFF}}c |
				>{\columncolor[HTML]{FFFFFF}}c 
				>{\columncolor[HTML]{FFFFFF}}c }
			\hline
			Type                                                               & Network                                                     & \multicolumn{2}{c|}{\cellcolor[HTML]{FFFFFF}Brain US} & \multicolumn{2}{c|}{\cellcolor[HTML]{FFFFFF}GlaS} & \multicolumn{2}{c}{\cellcolor[HTML]{FFFFFF}MoNuSeg} \\ \hline
			&                                                             & F1                         & IoU                        & F1                   & IoU                  & F1                    & IoU                   \\ \cline{3-8} 
			& FCN \cite{badrinarayanan2017segnet}                                                         & 82.79                           & 75.02                       & 66.61                     & 50.84                 & 28.84                      & 28.71                  \\
			\begin{tabular}[c]{@{}c@{}}Convolutional \\ Baselines\end{tabular} & U-Net \cite{ronneberger2015u}                                                       & 85.37                           & 79.31                       & 77.78                     & 65.34                 & 79.43                      & 65.99                  \\
			& U-Net++ \cite{zhou2018unet++}                                                    & 86.59                           & 79.95                       & 78.03                          & 65.55                      & 79.49                           & 66.04                       \\
			& Res-UNet \cite{xiao2018weighted}                                                   & 87.50                            & 79.61                       & 78.83                     & 65.95                 & 79.49                      & 66.07                  \\ \hline
			\begin{tabular}[c]{@{}c@{}}Fully Attention \\ Baseline\end{tabular}    & \begin{tabular}[c]{@{}c@{}}Axial Attention\\ U-Net \cite{wang2020axial}\end{tabular} & 87.92                           & 80.14                       & 76.26                     & 63.03                 & 76.83                      & 62.49                  \\ \hline
			& Gated Axial Attn.                                           & 88.39                           & 80.7                        & 79.91                     & 67.85                 & 76.44                      & 62.01                  \\
			Proposed                                                           & LoGo                                                        & 88.54                           & 80.84                       & 79.68                     & 67.69                 & 79.56                      & 66.17                  \\
			& MedT                                                        & \textbf{88.84}                           & \textbf{81.34}                       & \textbf{81.02}                     & \textbf{69.61}                 & \textbf{79.55}                      & \textbf{66.17}    \\ \hline             
		\end{tabular}
	}
\vskip -20 pt
	\end{table}
	
	For quantitative analysis, we use F1 and IoU scores for comparison. The quantitative results are tabulated in Table \ref{table:res}. It can be noted that for datasets with relatively more images like Brain US, fully attention (transformer) based baseline performs better than convolutional baselines. For GlaS and MoNuSeg datasets, convolutional baselines perform better than fully attention baselines as it is difficult to train fully attention models with less data \cite{dosovitskiy2020image}. The proposed method is able to overcome such issue with the help of gated axial attention and LoGo both individually perform better than the other methods. Our final architecture MedT performs better than Gated axial attention, LoGo and all the previous methods. The improvements over fully attention baselines are 0.92 \%,  4.76 \% and 2.72 \% for Brain US, GlaS and MoNuSeg datasets, respectively. Improvements over the best convolutional baseline are 1.32 \%, 2.19 \% and 0.06 \%. All of these values are in terms of F1 scores. For the ablation study, we use the Brain US
	data for all our experiments. The results for the same has been tabulated in Table \ref*{table:abl}.
	
\begin{figure}[!]
	\centering
	\includegraphics[width=0.8\linewidth]{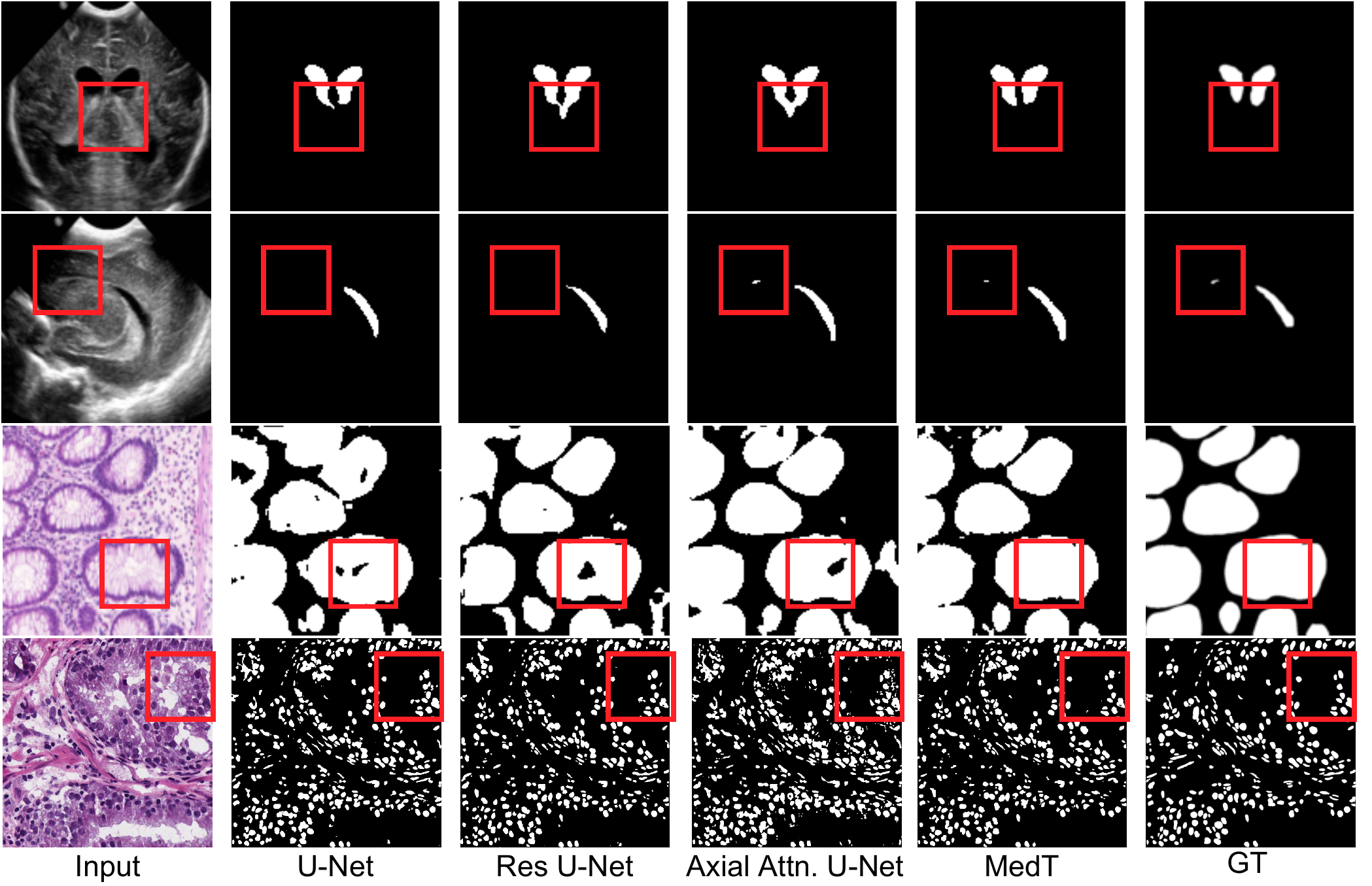}\\
	\vskip -10 pt
	\caption{Qualitative results on sample test images from Brain US, Glas and MoNuSeg datasets. The red box highlights regions where exactly MedT performs better than the other methods in comparison making better use of long range dependencies.}
	
	\label{fig:res}
	\vskip -35 pt
\end{figure}

	Furthermore, we visualize the predictions from U-Net \cite{ronneberger2015u}, Res-UNet \cite{xiao2018weighted}, Axial Attention U-Net \cite{wang2020axial} and our proposed method MedT in Fig \ref{fig:res}. It can be seen that the predictions of MedT captures the long range dependencies really well. For example, in the second row of Fig \ref{fig:res}, we can observe that the small segmentation mask highlighted on red box goes undetected in all the convolutional baselines. However, as fully attention model encodes long range dependencies, it learns to segment well thanks to the encoded global context. In the first and fourth row, other methods make false predictions at the highlighted regions as those pixels are in close proximity to the segmentation mask. As our method takes into account pixel-wise dependencies that are encoded with gating mechanism, it is able to learn those dependencies better than the axial attention U-Net. This makes our predictions more precise as they do not miss-classify pixels near the segmentation mask.   
	
	\begin{table}[]
		\vskip -20 pt
		\centering
		\caption{Ablation Study}
		\label{table:abl}
		\resizebox{0.85\linewidth}{!}{
			\begin{tabular}{c|c|c|c|c|c|c|c|c}
				\hline
				Network  & U-Net \cite{ronneberger2015u} & \begin{tabular}[c]{@{}c@{}}Res-UNet \cite{xiao2018weighted}\end{tabular} & \begin{tabular}[c]{@{}c@{}}Axial UNet \cite{wang2020axial}\end{tabular} & \begin{tabular}[c]{@{}c@{}}Gated Axial UNet\end{tabular} & \begin{tabular}[c]{@{}c@{}} Global\\ only\end{tabular} & \begin{tabular}[c]{@{}c@{}} Local\\ only\end{tabular} & LoGo  & MedT  \\ \hline
				F1 Score & 85.37 & 87.5                                                & 87.92                                                & 88.39                                                         & 87.67                                                         & 77.55                                                        & 88.54 & 88.84
			\end{tabular}
		}
		\vskip -30 pt
	\end{table}

	\section{Conclusion}
	
	In this work, we explored the use of transformer-based architectures for medical image segmentation. Specifically, we propose a gated axial attention layer which is used as the building block for multi-head attention models. We also proposed a LoGo training strategy to train the image in both full resolution as well in patches. The global branch helps learn global context features by modeling long-range dependencies, where as the local branch focus on finer features by operating on patches. Using these, we propose MedT (Medical Transformer) which has gated axial attention as its main building block for the encoder and uses LoGo strategy for training. Unlike other transformer-based model the proposed method does not require pre-training on large-scale datasets. Finally, we conduct extensive experiments on three datasets where we achieve a good performance for MedT over ConvNets and other related transformer-based architectures.
	
	\section*{Acknowledgment}
	This work was supported by the NSF grant 1910141.
	
		\bibliographystyle{splncs04}
	\bibliography{paper0116}
	\title{Supplementary Material for Medical Transformer: Gated Axial-Attention for Medical Image Segmentation}
	%
	%
\author{Jeya Maria Jose Valanarasu\inst{1} \and 
	Poojan Oza\inst{1} \and
	Ilker Hacihaliloglu\inst{2} \and
	Vishal M. Patel\inst{1}
}
%
\authorrunning{JMJ Valanarasu et al.}
%
\institute{Johns Hopkins University, Baltimore, MD, USA \and
	Rutgers, The State University of New Jersey, NJ, USA 
}
	\maketitle              

	In this supplementary material, we describe more details about the datasets that we used; provide more intricate details on our proposed architecture and training strategy; conduct an ablation study for our proposed methods; conduct an analysis on the number of parameters and present some more results.
	
	\section{Dataset details}
	
	In this section, we describe the datasets that we use in this paper in detail.
	\subsection{Brain US Dataset} Intraventricular hemorrhage (IVH) which results in the enlargement of brain ventricles is one of the main causes of preterm brain injury. The main imaging modality used for diagnosis of brain disorders in preterm neonates is cranial US because of its safety and cost-effectiveness. Also, absence of septum pellucidum is an important biomarker for septo-optic dysplasia diagnosis. Automatic segmentation of brain ventricles and septum pellucidum from these US scans is essential for accurate diagnosis and prognosis of these ailments. After obtaining institutional review board (IRB) approval, US scans were collected from 20 different premature neonates (age $<$ 1 year). The total number of images collected were 1629 with annotations out of which 1300 were allocated for training and 329 for testing. We resize the images to $128\times128$ for all our experiments.

	\subsection{GLAS Dataset} GLAnd Segmentation (GLAS) datatset \cite{sirinukunwattana2017gland} contains microscopic images of Hematoxylin and Eosin (H\&E) stained slides and the corresponding ground truth annotations by expert pathologists. It contains a total of 165 images which are split into 85 images for training and 80 for testing. Since the images in the dataset are of different sizes, we resize every image to a resolution of $128\times128$ for all our experiments. 
	
	\subsection{MoNuSeg Dataset} MoNuSeg dataset \cite{kumar2019multi,kumar2017dataset} was created using H\&E stained tissue images captured at 40x magnification. This dataset is diverse as it contains images across multiple organs and patients. The training data contains 30 images with around 22000 nuclear boundary annotations. The test data contains 14 images which have over 7000 nuclear boundary annotations. We resize the images to $512\times512$ for all our experiments.

	
	\section{MedT details}
	
	Medical Transformer (MedT) uses gated axial attention layer as the basic building block and uses  LoGo strategy for training. MedT has two branches - a global branch and local branch. The input to both of these branches are the feature maps extracted from an initial conv block. This block has 3 conv layers, each followed by a batch normalization and ReLU activation. In the encoder of both branches, we use our proposed transformer layer while in the decoder, we use a conv block. The encoder bottleneck contains a $1 \times 1$ conv layer followed by normalization and two layers of multi-head attention layers where one operates along height axis and the other along width axis. Each multi-head attention block is made up of the proposed gated axial attention layer. Note that each multi-head attention block has 8 gated axial attention heads. The output from the multi-head attention blocks are concatenated and passed through another $1 \times 1$ conv which are added to residual input maps to produce the output attention maps. In each decoder block, we have a conv layer followed by an upsampling layer and ReLU activation. We also have skip connections between each encoder and decoder blocks in both the branches.
	
	In the global branch of MedT, we have 2 blocks of encoder and 2 blocks of decoder. In the local branch, we have 5 blocks of encoder and 5 blocks of decoder.  
	
	\section{Training details}
	
	We use a batch size of 4, Adam optimizer \cite{kingma2014adam} and a learning rate of 0.001 for our experiments. The network is trained for 400 epochs. While training the gated axial attention layer, we do not activate the training of the gates for the first 10 epochs. We use a Nvidia Quadro 8000 GPU for all our experiments. 
	
	\section{Analysis}
	In this section, we present an analysis over some of the parameters and methods we used for our proposed method.
	
	\subsection{Ablation Study}
	
	For the ablation study, we use the Brain US data for all our experiments. We first start with a standard U-Net. Then, we add residual connections to the U-Net making it a Res-UNet. Now, we replace all the convolutional layers in the encoder of Res-UNet with axial attention layers. This configuration is Axial Attention UNet inspired from \cite{wang2020axial}. Note that in this configuration we have an additional conv block at the front for feature extraction. Next, we replace all the axial attention layers from the previous configuration with gated axial attention layers. This configuration is denoted as Gated Axial attention. We then experiment using only the global branch and local branch individually from LoGo strategy. This shows that using just 2 layers in the global branch is enough to get a decent performance. The local branch in this configuration is tested on the patches extracted from the image. Then, we combine both the branches to train the network in an end-to-end fashion which is denoted as LoGo. Note that in this configuration the attention layers used are just axial attention layers \cite{wang2020axial}. Finally, we replace the axial attention layers in LoGo with gated axial attention layers which leads to MedT. The ablation study shows that each individual components of MedT provides useful contribution to improve the performance. 
	
	\begin{table}[]
		\centering
		\caption{Ablation Study}
		\begin{tabular}{c|c|c|c|c|c|c|c|c}
			\hline
			Network  & U-Net \cite{ronneberger2015u} & \begin{tabular}[c]{@{}c@{}}Res-\\ UNet \cite{xiao2018weighted}\end{tabular} & \begin{tabular}[c]{@{}c@{}}Axial\\ UNet \cite{wang2020axial}\end{tabular} & \begin{tabular}[c]{@{}c@{}}Gated \\ Axial\\ UNet\end{tabular} & \begin{tabular}[c]{@{}c@{}} Global\\ only\end{tabular} & \begin{tabular}[c]{@{}c@{}} Local\\ only\end{tabular} & LoGo  & MedT  \\ \hline
			F1 Score & 85.37 & 87.5                                                & 87.92                                                & 88.39                                                         & 87.67                                                         & 77.55                                                        & 88.54 & 88.84
		\end{tabular}
	\end{table}
	
	\subsection{Number of Parameters}
	
	\begin{table}[]
		\centering
		\caption{Comparison in     terms of number of parameters between the proposed method with the existing methods.}
		\label{table:num_parameters}
		\begin{tabular}{c|c|c|c|c|c|c|c|c}
			\hline
			Network    & FCN \cite{badrinarayanan2017segnet}   & U-Net \cite{ronneberger2015u} & \begin{tabular}[c]{@{}c@{}}U-Net \cite{ronneberger2015u}\\ (mod)\end{tabular} & \begin{tabular}[c]{@{}c@{}}Res-\\ UNet \cite{xiao2018weighted}\end{tabular} & \begin{tabular}[c]{@{}c@{}}Res\\ UNet \cite{xiao2018weighted}\\ (mod)\end{tabular} & \begin{tabular}[c]{@{}c@{}}Axial \\ UNet \cite{wang2020axial}\end{tabular} & \begin{tabular}[c]{@{}c@{}}Gated \\ Axial\\  UNet\end{tabular} & MedT  \\ \hline
			Parameters & 12.5 M & 3.13 M & 1.3 M                                                   & 5.32 M                                              & 1.34 M                                                      & 1.3 M                                                 & {1.3} M                                                          & 1.4 M \\ \hline
			F1 Score   & 82.79  & 87.71  & 85.37                                                   & 87.73                                               & 87.5                                                        & 87.92                                                 & 88.39                                                          & \textbf{88.84}
		\end{tabular}
	\end{table}
	
	Although MedT is a multi-branch network, we reduce the number of parameters by using only 2 layers of encoder and decoder in the global branch and making the local branch operate on only patches of image. Also, the proposed gated axial attention block adds only 4 more learnable parameters to the layer. In Table~\ref{table:num_parameters}, we compare the number of parameters with other methods. U-Net corresponds to the original implementation according to \cite{ronneberger2015u}. U-Net (mod) corresponds to the U-Net configuration with reduced number of filters so as to match the number of parameters in MedT. Similarly, Res-UNet and Res-UNet (mod) corresponds to configurations with more and less number of parameters by adjusting the number of filters. We do this to show that even with more number of parameters, the baselines do not exceed MedT in terms of performance indicating that the improvement is not due to slight change in the number of parameters.

	\section{Results}
	
	\begin{figure}[]
		\centering
		\includegraphics[width=1\linewidth]{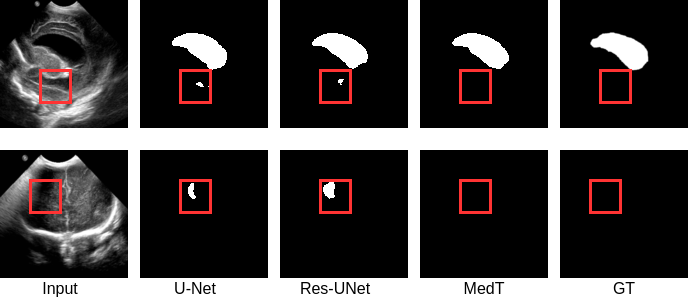}\\
		
		\caption{Qualitative Results. The red box highlights the regions where our proposed method outperforms the convolutional baselines. }
		
		\label{qual}
	\end{figure}
	
	We present some additional qualitative results on top of the qualitative results presented in the main paper. In Fig \ref{qual}, we visualize the predictions for our proposed method MedT along with the predictions for baselines UNet and Res-UNet for a couple of US scans. In both the samples, it can be seen that the regions that are highlighted in the red box are miss-classified to be brain ventricles for the convolutional baselines. However, our proposed attention based MedT does not make the same mistake.

	\section{Concurrent works}
	Very recently, TransUNet \cite{chen2021transunet} was proposed which uses a transformer-based encoder operating on sequences of image patches and a convolutional decoder with skip connections for medical image segmentation. As TransUNet is inspired by ViT, it is still
	dependent on pretrained weights obtained by training on a large image corpus. TransFuse \cite{zhang2021transfuse} was recently proposed for polyp segmentation tasks using a parallel CNN branch and transformer branch fused using a BiFusion module. Unlike these works, we explore the feasibility of applying transformers working on only self-attention mechanisms as an encoder for medical image segmentation and without any need for pre-training.
	
\bibliographystyle{splncs04}
\bibliography{medt}
\end{document}